\documentclass[letterpaper]{article} 
\usepackage{aaai24}  
\usepackage{latexsym}
\usepackage{float}
\usepackage{color}
\usepackage{amssymb}
\usepackage{dsfont}
\usepackage{amsmath}
\usepackage{algorithm}
\usepackage{algorithmicx}
\usepackage[noend]{algpseudocode}
\usepackage{booktabs}
\usepackage{times}  
\usepackage{helvet}  
\usepackage{courier}  
\usepackage[hyphens]{url}  
\usepackage{graphicx} 
\usepackage{comment}
\usepackage{natbib}  
\usepackage{caption} 
\usepackage{newfloat}
\usepackage{listings}
\DeclareCaptionStyle{ruled}{labelfont=normalfont,labelsep=colon,strut=off} 
\floatstyle{ruled}
\newfloat{listing}{tb}{lst}{}
\floatname{listing}{Listing}
\nocopyright

\begin{document}

\renewcommand{\algorithmicrequire}{\textbf{Input:}} 
\renewcommand{\algorithmicensure}{\textbf{Initialize:}} 
\newcommand{\Q}{\mathcal{Q}}
\newcommand{\F}{\mathcal{F}}
\renewcommand{\P}{\mathcal{P}}
\renewcommand{\L}{\mathcal{L}}
\newcommand{\R}{\mathcal{R}}
\newcommand{\RR}{\mathbb{R}}
\newcommand{\bv}[1]{\mathbf{#1}}
\newcommand{\A}{\mathcal{A}}
\newtheorem{theorem}{Theorem}
\newtheorem{remark}{Remark}
\newtheorem{proof}{Proof}
\newtheorem{proposition}[theorem]{Proposition}
\newtheorem{corollary}[theorem]{Corollary}
\newtheorem{definition}{Definition}
\newtheorem{example}{Example}
\newcommand{\eqnref}[1]{Eqn.~\ref{#1}}
\newcommand{\secref}[1]{Section~\ref{#1}}
\newcommand{\appref}[1]{Appendix~\ref{#1}}
\newcommand{\figref}[1]{Fig.~\ref{#1}}
\newcommand{\thmref}[1]{Theorem~\ref{#1}}
\newcommand{\lemref}[1]{Lemma~\ref{#1}}
\newcommand{\propref}[1]{Proposition~\ref{#1}}
\newcommand{\I}{\mathcal{I}}
\newcommand{\RB}[1]{{\color{blue}\textbf{RB:} #1}}
\newcommand{\LL}[1]{{\color{blue}#1}}
\newcommand{\LLR}[1]{{\color{red}#1}}
\newcommand{\wen}[1]{{\color{green}#1}}
\renewcommand{\S}{\mathcal{S}}
\newcommand{\M}{\mathcal{M}}
\newcommand\ind[1]{\ensuremath{\mathds{1}\left[#1\right]}}
\newcommand{\E}[2]{\mathbf{E}_{#1}\left[#2\right]}
\newcommand{\N}{\mathcal{N}}
\newcommand{\interior}[1]{%
  {\kern0pt#1}^{\mathrm{o}}%
}


\title{BadRL: Sparse Targeted Backdoor Attack Against Reinforcement Learning}
\author {
    Jing Cui\textsuperscript{\rm 1},
    Yufei Han\textsuperscript{\rm 2},
    Yuzhe Ma\textsuperscript{\rm 3},
    Jianbin Jiao\textsuperscript{\rm 1},
    Junge Zhang\textsuperscript{\rm 4,1}\thanks{Corresponding author}
}
\affiliations {
    \textsuperscript{\rm 1}University of Chinese Academy of Sciences,
    \textsuperscript{\rm 2}INRIA,
    \textsuperscript{\rm 3}Microsoft Azure AI,\\
    \textsuperscript{\rm 4}Institute of Automation, Chinese Academy of Sciences\\
    cuijing21@mails.ucas.ac.cn, yufei.han@inria.fr, yuzhema@microsoft.com, \\jiaojb@ucas.ac.cn,  jgzhang@nlpr.ia.ac.cn
}
\maketitle

\begin{abstract}
Backdoor attacks in reinforcement learning (RL) have previously employed intense attack strategies to ensure attack success. However, these methods suffer from high attack costs and increased detectability. In this work, we propose a novel approach, BadRL, which focuses on conducting highly sparse backdoor poisoning efforts during training and testing while maintaining successful attacks. Our algorithm, BadRL, strategically chooses state observations with high attack values to inject triggers during training and testing, thereby reducing the chances of detection. In contrast to the previous methods that utilize sample-agnostic trigger patterns, BadRL dynamically generates distinct trigger patterns based on targeted state observations, thereby enhancing its effectiveness. Theoretical analysis shows that the targeted backdoor attack is always viable and remains stealthy under specific assumptions. Empirical results on various classic RL tasks illustrate that BadRL can substantially degrade the performance of a victim agent with minimal poisoning efforts (\textbf{0.003\%} of total training steps) during training and infrequent attacks during testing.

    
\end{abstract}


\vspace{-2mm}
\section{Introduction}

Prior works have demonstrated that reinforcement learning (RL) is susceptible to backdoor poisoning attacks~\cite{TrojRL,BACRL,TPBA,CRNN,ppt,JHT}. Similar to Supervised Learning, a backdoor attack in RL pursues dual objectives~\cite{TrojRL}. On the one hand, a backdoored RL agent trained with poisoned training data should perform comparably to an adversary-free policy when state observations do not contain the adversary-designed trigger signal. On the other hand, once the trigger signal is injected into the state observations, the backdoored policy should significantly degrade the agent's performance. However, backdoor attacks in RL can be more hazardous due to the sequential nature of RL. An incorrect action triggered by an attack may reduce the immediate reward and guide the agent to a low-value state (e.g., a failure state in computer games), resulting in a small cumulative future reward after a series of actions.

This sequential characteristic of RL poses unique challenges for organizing backdoor poisoning attacks in RL systems. \textbf{During the training of the backdoored policy}, the adversary must comprehend the state values and corresponding optimal actions to consider the entire future of the agent, i.e., the actions, state observations, and rewards it may receive in the future, to organize the backdoor policy training process. Specifically, in each round of policy learning, the adversary must decide which state observations to inject the trigger signals and modify the corresponding rewards and/or actions to minimize future rewards instead of only minimizing the instant rewards. Similarly, \textbf{during testing time}, when the backdoored policy is deployed, the adversary needs to strategically choose state observations of the agent to embed the trigger and enhance the degradation impact throughout the agent's course of action. 

Previous studies ~\cite{TrojRL,TPBA} employed an intense attack strategy during training and testing time to ensure the success of backdoor attacks. However, those attack methods do not consider the impact of each trigger-injection operation on future rewards. Performing such intense attacks during training time can lead to unnecessarily high attack costs with little additional impact. For example, in Breakout, targeting the initial game state, which may not directly impact success or failure, would be ineffective in causing the agent to fail. Additionally, intense attacks introduce excessive perturbations to the trained policy, leading to the degradation of the RL task even without the trigger signal. During testing time, frequent testing-time poisoning operations incur high costs and increase the noticeability of the attack.

Our study aims to address the bottlenecks of intense attack costs in prior works. We focus on designing an efficient backdoor attack strategy, which only spends highly sparse backdoor poisoning efforts at training and testing time yet delivers successful attacks. Specifically, at the training stage, the attacker aims to craft the cells of state observations, actions and received reward signals up to a certain fraction of the total time steps in the training process to learn the backdoored policy (see the attack formulation in Eq.4 and 8), and at the testing stage, once the backdoored policy is trained and deployed in the testing phase, the attacker is constrained to inject the trigger into as few time steps during testing to deliver the attack. Compared to previous works \cite{TrojRL}, where the trigger is injected into consecutive time steps during testing until the attack goal is achieved, our testing-time trigger injection effort has a low attack frequency.

Our approach results in sparse attacks from two perspectives.
\textbf{Firstly}, we adopt a sample-specific approach to generate backdoor triggers so that those triggers become easier to learn and less susceptible to forgetting when the policy model is updated with poison-free samples. The generation process aims at maximizing the mutual information between the gradients of the policy model with respect to trigger-embedded and clean input states. Dragging two gradients close align the backdoor poisoning task with the main learning task during training. Consequently, while the agent learns the main task, it also strives to consolidate the association between the backdoor trigger and the attacker-desired action. Also, such triggers could evade state-of-the-art defense. 
In contrast, ~\cite{TrojRL,TPBA,CRNN,ppt,JHT} manually choose the trigger signal in a sample-agnostic way, which is often easy to detect and mitigate by existing defenses. Additionally, manually designed triggers may introduce unexpected artifacts into targeted states, making them difficult for the victim model to memorize. Hence, those triggers result in intense poisoning efforts during training to forge the backdoor mapping, which may deform the agent's performance in adversary-free environments and increase the overall attack cost. \textbf{Secondly}, we propose to assess the impact of poisoning a given time step on future rewards, namely evaluating the attack values of backdoor poisoning. Our backdoor attack strategy involves selecting only the states with high attack values for backdoor poisoning, significantly reducing the attack frequency. By combining sample-specific triggers and critical state selection, our proposed backdoor attack can reduce the proportion of poisoned samples to only \textbf{0.003\%} of the total training samples during the RL training process, which is 1/10 of state-of-the-art baseline \cite{TrojRL} (\textbf{0.025\%}) while achieving higher attack success rates. For instance, in the Breakout game, our attack strategy achieves a three times higher attack success rate without affecting the performance of the poison-free learning task, yet requiring only 1/2 the testing-time attack frequency compared to the baseline (Shown in Table \ref{table:overall performance}). 

Our main contributions are summarized in three folds.

We propose BadRL, a novel targeted backdoor poisoning attack against reinforcement learning algorithms. Unlike existing backdoor attack algorithms that intensively inject attacks, BadRL adopts sparse trigger injection during training and testing to reduce the RL agent's overall performance (i.e., cumulative reward). Additionally, we provide theoretical analyses on the feasibility of backdoor poisoning attacks, demonstrating the existence of effective yet stealthy attacks under certain assumptions.
  
The proposed BadRL attack tackles the challenge of determining ``when to attack" to deliver sparse testing-time poisoning. Specifically, BadRL identifies a small subset of high attack-value states and performs sparse poisoning only on these selected states. Through experiments, we demonstrate that BadRL successfully conducts \emph{sparse backdoor poisoning efforts} during testing time, effectively undermining the performance of the RL agent. 

BadRL adopts a sample-specific trigger design using mutual information, which is difficult to detect using state-of-the-art countermeasures and meets a tight budget constraint, resulting in more cost-effective poisoning efforts during training. Our approach achieves an almost 100\% success rate, requiring only 1/10 of the poisoning efforts compared to the state-of-the-art methods.

\section{Related Work}

\textbf{Adversarial attack in RL.} 
Adversarial attacks in Reinforcement Learning(RL) have been explored in several works.
\cite{DBLP:conf/iclr/SunHH21} propose a general adversarial attack approach by measuring the policy divergence resulting from poisoning a trajectory to make an attack decision. \cite{adrl} demonstrates that for deep RL problems, one can construct adversarial examples without requiring them to be superior to the best opponent's policy. Moreover, \cite{rewardpoisoning} and \cite{rakhsha2020policyteaching} present studies where the victim policy can be manipulated to converge to an attacker-desired policy by modifying the reward or transition function.

\textbf{Backdoor attack against RL.}
As defined in \cite{LiTNNLS2022}, backdoor attacks are a family of training-time targeted attack against ML systems. They bypass DNN decision-making, activating hidden backdoors for compromised behavior. \cite{gu2019badnets} demonstrates that backdoor attacks can be accomplished by introducing trigger patterns into the training data. Their work laid the foundation for optimizing backdoor attacks, and subsequent studies, such as \cite{saha2019hiddentrigger} and \cite{hiber2022back}, have built upon this research. 
Various studies have explored data-poisoning-based backdoor attacks, as evidenced by the works of \cite{saha2022datapoisoningbased} and \cite{carlini2022datapoisoningbased}. Additionally, \cite{jia2021modelpoisoningbased} investigates model poisoning backdoor attacks, \cite{liu2018trojaning} studies Trojan backdoor attacks, and \cite{adi2018watermarking} focuses on watermarking attacks. Our study focuses on backdoor attacks targeting online RL systems, where the adversary manipulates the training data to create a backdoored policy model. This policy executes attacker-desired actions, minimizing future rewards when an attacker-designated trigger signal is present in the victim agent’s state; otherwise, it usually functions. 
TrojDRL~\cite{TrojRL} is among the first attempts at backdoor attacks against RL using the untargeted threat model. They find that the trigger-to-action mapping can be established by injecting the trigger uniformly during training. During testing, consecutive attacks can lead to the destruction of the learning model. Prior works like \cite{BACRL,TPBA} select infected states using hand-crafted rules. For example, \cite{BACRL} studies backdoor attacks in competitive RL, selecting one of the opponent's actions as a trigger to switch the victim agent's policy to a fast-falling one.
\cite{TPBA} studies Partially Observable MDP (POMDP), which hides the trigger pattern in a sequence of input states and continuously manipulates the rewards during the trigger appearance duration. \cite{gong2022mind} studies the backdoor attack in the setting of offline RL, where attackers could either flip the reward signal for consecutively N time steps or uniformly sampled $N$ time steps. The consecutive reward manipulation could drift policy distribution severely and reduces the attack stealthiness. In contrast, our approach prioritizes sparse poisoning efforts, aiming to affect as few
training steps in the training episodes as possible. We choose a state to infect by estimating the attack value on the concerned data and restrict the modification ability to single-step information only. 

We emphasize that our contribution differs from those in \cite{BACRL,TPBA,gong2022mind} in the following perspectives. \cite{BACRL} focuses on a competitive two-player game with RL agents, whereas our research centers on single-agent RL tasks. The distinction lies in two key aspects. Firstly, in the competitive game,
the backdoor attack is triggered when the victim agent observes
a certain action her opponent performs. Secondly, the training process for the backdoor policy must consider interactions among the victim agent, the environment, and the opponent agent’s actions simultaneously. In our case, the backdoor attack activates when the victim agent observes state data containing the desired trigger pattern set by the attacker. Additionally, learning the victim’s backdoor policy only involves interactions with the environment, excluding considerations of opponent actions. 
In the work by \cite{TPBA}, the trigger is set as a specific sequential pattern of state observations, unlike our study, where individual state observations embedded with the trigger are considered. Additionally, the adversary in their research spreads the poisoning efforts across consecutive time steps. Conversely, our threat model limits the attacker to strategically select as few time steps as possible to inject poisoned training data. 
\cite{gong2022mind} focuses on offline RL backdoor attacks, while ours targets online RL. Offline RL relies solely on previously collected training data without interaction with the environment. Therefore, poisoning the current time step does not impact the distribution of future training data. Online RL, however, generates training samples via agent-environment interactions. Poisoning the current step can have a profound impact on the distribution of future training samples of the victim agent, demanding meticulous control for successful backdoor attacks.

In contrast to \cite{adrl,rewardpoisoning,rakhsha2020policyteaching}, our proposed method focuses on sparse and targeted backdoor poisoning against RL. We aim to consistently favor an attacker-desired action when a trigger is present in the state observation, while resembling a normal policy in backdoor-free testing conditions. Data poisoning attacks against RL, on the other hand, undermine policy performance globally, resulting in abnormal behaviors during testing. As a result, the backdoor attack is more evasive than the data poisoning attack against RL systems. The threat model for backdoor attacks against RL differs from traditional non-RL tasks like classification. In RL, attack success cannot be solely measured by the attack success rate; it must consider the cumulative reward degradation factor. The attacker must optimize the attack budget to maximize the reduction in accumulated rewards within the given limit. In contrast, classification tasks have no such considerations, as the output of one step does not affect subsequent steps.

\vspace{-1.5mm}
\section{Preliminary}

The underlying environment is a Markov Decision Process (MDP) $\M=(\S, \A, \R, \P, \mu_0)$, where $\S$ is the original state space, $\mathcal{A}$ is the action space, $\mathcal{R}: \S \times \A \mapsto \RR$ is the reward function, $\P: \S \times \A \mapsto \Delta (\S)$ is the transition model ($\Delta(\S)$ is a distribution over $\S$), and $\mu_0$ is the initial state distribution. 
At each round $t$, let $s_t\in \S$ denote the state of the environment and $a_t\in \A$ denotes the chosen action.
A policy is a function $\pi:\S\mapsto \Delta(\A)$ that maps any state $s$ to a distribution over actions. The value function of a policy $\pi$ with respect to an initial state $s$ is defined as the cumulative reward obtained by the agent, starting from state $s$ and the following policy $\pi$ in all future rounds, i.e.,
\begin{small}
\begin{equation}
    V^\pi(s)=\E{}{\sum_{t=0}^T \R(s_t, a_t)\mid \pi, s_0=s}, \forall s\in \S,
\end{equation}
\end{small}
where $T$ is the total number of rounds.
The learning objective of an RL agent is to find the optimal policy $\pi^*$ that attains the maximum value:
\begin{small}
\begin{equation}
    \pi^* = \arg\max_{\pi} \;\, \E{s_0\sim \mu_0}{V^\pi(s_0)}.
\end{equation}
\end{small}
The state-action value function is defined as:
\begin{small}
\begin{equation}\label{eq:Q}
    Q^\pi(s, a)=\E{}{\sum_{t=0}^T \R(s_t, a_t)\mid \pi, s_0=s, a_0=a}.
\end{equation}
\end{small}
$Q^*$ denotes the state-action value of the optimal policy $\pi^*$.

\section{Threat Model of Targeted Backdoor Attack}
\textbf{Attacker Knowledge.} We adopt a black-box attack, which means the attacker does \textit{not} know the RL algorithm used by the victim agent. The attacker does \textit{not} know the underlying clean MDP environment either. However, before the attack happens, we assume the attacker has access to a simulator to interact with the clean environment for an arbitrary number of rounds, which has been widely used in prior works, e.g.,~\cite{zhang2020adaptive}. For example, an attacker against autonomous driving systems may use his own driving facilities to collect trigger-free driving data inside similar areas as the target autonomous driving agent. With the simulator, the attacker can obtain accurate estimates of relevant statistics of the underlying MDP. In particular, the attacker can obtain an estimate of the optimal state-action value function $\tilde Q^*(s, a)$, which is used to select high attack-value states during the attack. 

\textbf{Attacker Ability.} Let ($s_t$, $a_t$, $r_t$) denote the poison-free data and ($\tilde s_t$, $\tilde a_t$, $\tilde r_t$) denote the poisoned counterparts. The attacker can perturb both training and testing data during the online interaction between the victim RL agent and the environment. However, the perturbation is strictly limited to current step information only. Specifically,

\textbf{(1)} In the training phase of backdoor attack, let $s_t$ denote the original state of the agent at round $t$.
The attacker can inject a trigger $\delta$ into the state $s_t$ to make the agent perceive a trigger-embedded state $\tilde s_t = s_t + \delta$. Then the agent selects an action $a_t$ based on $\tilde s_t$. Similar to prior works~\cite{TrojRL,liu2021provably}, we assume our attacker can override the selected action and force the agent to take a different action $\tilde a_t$, which is formalized as the \emph{strong attacker}; we also consider the \emph{weak attacker} case where an attacker has no rights to modify action and the agent action $a_t$ stays. Then the environment generates the reward $r_t$ according to the true state $s_t$ and the attacker-modified action $\tilde a_t$ (or action $a_t$ in the case of the weak attacker). The attacker can further modify the reward $r_t$ to $\tilde r_t$. At the end of the round, the agent observes the perturbed reward $\tilde r_t$ and transits to the next state $s_{t+1}$ according to $s_t$ and $\tilde a_t$(or $a_t$ for a weak attacker). In summary, the agent observes poisoned data point $(\tilde s_t$, $\tilde a_t (a_t)$, $\tilde r_t)$ at round $t$ of training. Although the attacker can arbitrarily perturb each data point, we restrict the power of the attacker to only poison a small fraction of the training data, i.e., 
\vspace{-1mm}
\begin{small}
\begin{equation}
    \sum_{t=1}^T \ind{(s_t, a_t, r_t)\neq (\tilde{s_t}, \tilde{a_t}(a_t), \tilde{r_t})}\le \epsilon T,
\end{equation}
\end{small}
where $\epsilon\ll 1$ is the attack budget and $T$ is the total rounds of training steps of the agent. Note that this constraint does not aim to provide a theoretical guarantee for evading visual inspection. However, it is set to encourage sparse attacks, in order to reduce the attack frequency. In practice, the learner may inspect state observations regularly to perform sanitary checks. Setting a low attack frequency (a highly sparse attack) can hence reduce the possibility of being flagged, which makes the attack stealthy. 

\textbf{(2)} In the testing phase, the attacker has rather limited ability compared to the training phase. The attacker can only perturb the state perceived by the agent to $\tilde s_t=s_t+\delta$. As a result, the agent selects an action $a_t$ according to the perturbed state $\tilde s_t$ and follows the backdoor policy. Both the reward $r_t$ and the next state $s_{t+1}$ are generated according to $s_t$ and $a_t$ following the clean MDP environment. 

\textbf{Attack Goal.} To characterize our attack goal, we first define the value function $\tilde V^\pi(s)$ of a policy function $\pi$.  
$\tilde V^\pi(s)$ is the cumulative reward that the victim agent can obtain under backdoor attacks following the policy $\pi$ during the testing phase:
\vspace{-1mm}
\begin{small}
\begin{equation}
\tilde V^\pi(s)=\E{}{\sum_{t=0}^T \R(\tilde s_t, a_t)\mid \pi, s_0=s},
\end{equation}
\end{small}
where $\tilde s_t$ is the state observation received by the victim agent at the time step $t$ of the testing phase. $a_t\sim \pi(\tilde s_t)$. Note that $\tilde s_t=s_t$ if $s_t$ is not embedded with any trigger signal. Otherwise, $\tilde s_t=s_t+\delta$.
We consider targeted backdoor poisoning, in which the attacker has a set of targeted states $\S^\dagger\subset \S$ and a target action $a^\dagger$.  
The attacker aims at forcing the agent into learning a sub-optimal policy $\tilde \pi$ (the backdoor policy) during the training phase such that the expected value of $\tilde \pi$ is minimized, i.e. $\E{s_0\sim\mu_0}{\tilde V^{\tilde \pi(s)}}$ is minimized as much as possible. 
Meanwhile, the attacker desires the following two properties on the backdoor policy $\tilde \pi(s)$:
\vspace{-1.5mm}
\begin{small}
\begin{equation}\label{eq:desire_1}
\tilde \pi(s+\delta) = a^\dagger,\forall s\in \S^\dagger,
\end{equation}
\end{small}
\begin{small}
\begin{equation}\label{eq:desire_2}
\E{s_0\sim \mu_0}{V^{\tilde \pi}(s_0)}=\E{s_0\sim \mu_0}{V^{\pi^*}(s_0)},
\end{equation}
\end{small}
where $\pi^*$ is the optimal policy with respect to the clean environment $\M$. We now explain the attack goals in detail. During the testing phase, the poisoned policy $\tilde \pi$ is executed by the agent. First, as characterized in Equation~\eqref{eq:desire_1}, the attacker desires that when the agent encounters any targeted state $s\in \S^\dagger$ during the testing phase, injecting the trigger into the state $s$ will mislead the agent to choose some target action $a^\dagger$ following the backdoor policy. The attacker chooses the target action to minimize the cumulative reward of the agent. This is a standard attack goal in common backdoor attacks. 
Second, Equation~\eqref{eq:desire_2} indicates that if no trigger is present, the attacker expects the backdoor policy to retain the performance of the optimal policy in a clean environment. This ensures that the backdoor policy behaves the same as the clean policy when no attack happens, making backdoor attacks more stealthy and less likely to be detected. We emphasize that the target action is designated by the attacker and remains fixed throughout an entire environment. This is consistent with the backdoor attack definition. The target action is chosen based on the advantage discrepancy between the targeted and the original action. More advantage discrepancies indicate stronger attack impacts

We highlight the difference in threat models between our work and TrojDRL~\cite{TrojRL}. In the threat model of TrojDRL, the attacker launches attacks consecutively at the testing phase until the RL agent completely fails the task. In contrast, our attack performs \textbf{sparse} attacks. 

\textbf{Attack Formulation.} Given the above threat model, the backdoor attack during the training phase can be formulated as the following optimization problem:
\begin{small}
\begin{equation}\label{attack:optimization}
\begin{aligned}
\min_{\tilde s_{1:T}, \tilde a_{1:T}, \tilde r_{1:T}}\quad &\E{s_0\sim\mu_0}{\tilde V^{\tilde \pi_T}(s_0)}\\
\text{s.t. \quad} &\sum_{t=1}^T \ind{(s_t, a_t, r_t)\neq (\tilde s_t, \tilde r_t, \tilde a_t(a_t)}\le \epsilon T,\\
&\tilde\pi_T(s+\delta) = a^\dagger,\forall s\in \S^\dagger,\\
&\E{s_0\sim \mu_0}{V^{\tilde \pi_T}(s_0)}=\E{s_0\sim \mu_0}{V^{\pi^*}(s_0)}.\\
\end{aligned}
\end{equation}
\end{small}

The main difficulty of solving Equation~\eqref{attack:optimization} lies in the first constraint. Since the attacker can only poison at most $\epsilon$ fraction of the training data, a decision is required at each step $t$ to decide whether the current data point $(s_t, a_t, r_t)$  should be manipulated, i.e., the problem of \textbf{when to attack}. 
Besides, once the attacker decides to poison the current data point, the key question for the attacker is how to produce the corresponding poisoned data point $(\tilde s_t, \tilde a_t(a_t), \tilde r_t)$ to maximize the attack effect, i.e., the problem of \textbf{how to attack}. In the following, we address both difficulties by proposing a \emph{sparse targeted backdoor attack} (BadRL) algorithm.

\setlength{\textfloatsep}{1mm}
\begin{algorithm}[t]
\begin{algorithmic}[1]
\footnotesize
\Require{Maximal training length $T$, target action $a^\dagger$, poisoning power $\eta$, trigger pattern $\delta$, poison percentage $k$, source action $a^\prime$, attack budget $\epsilon$} 
\Ensure{a queue of attack value $Q=\emptyset$, agent policy $\pi$, action space set $A \backslash \{a^\dagger\}$, $t_{attack} = 0$} 
\For{$t=1,...T$}     
    \State obtain the state $s_{t}$ obtained by the agent
    \State $\mathbf{attack}$ = False
    \If {$\frac{t_{attack}}{t} <\epsilon$}:
        \State $\mathbf{attack} = \mathbf{Poison}(s_{t},Q,k,a^\prime)$  $\backslash\backslash$ Algorithm~\ref{alg:A2}
    \EndIf
    \If {$\mathbf{attack}$}
            \State  $s_{t} = s_{t}+\delta$
            \State $t_{attack} = t_{attack} +1$
    \EndIf
    \State obtain the state $a_{t}$ with $s_{t}$
    \If {$\mathbf{attack}$}
        \If{$t$ is even}    
            \State $a_{t} = a^\dagger$
        \Else
            \State $a_{t}$ = arbitrary $a^\prime \in A$
        \EndIf
    \EndIf
    
    \State obtain the reward $r_{t}$, next state $s_{t+1}$
        \If{$\mathbf{attack}$}
            \If{$t$ is even}
            \State $r_t = \eta$
            \Else
            \State $r_t = -\eta$
            \EndIf
        \EndIf
    \State Update agent's policy $\pi$ with $(s_t, a_t, r_t)$
    \State $t=t+1$
\EndFor
\State \Return $\pi$

\end{algorithmic}
\caption{BadRL Algorithm.}
\label{alg:A1}
\end{algorithm}
\setlength{\floatsep}{1mm}

\begin{algorithm}[t]
\begin{algorithmic}[1]
\footnotesize
\Require{input state $s_{t}$, attack value queue $Q$, source action $a^\prime$, poison percentage $k$} 
\State compute action $a^{\pi^\dagger}=\pi^\dagger(s_{t})$ with attacker policy $\pi^\dagger$
\If{$a^{\pi^\dagger} == a^\prime$}  
        \State compute the attack value $v_{t}$
        \State append $v_{t}$ into $Q$
        \If {$v_{t}$ is no lower than $(1-k)\%$ of $Q$}
            \State  \Return= True
        \Else
            \State \Return= False
        \EndIf
\Else
\State \Return= False
\EndIf
\end{algorithmic}
\caption{$\mathbf{Poison}$ function}
\label{alg:A2}
\end{algorithm}

\section{BadRL Attack Framework}

\subsection{When to Attack: BadRL Specific Optimization}

Our BadRL attacker performs backdoor attacks only on the targeted states $\S^\dagger$. However, due to the budget constraint in Equation~\eqref{attack:optimization}, the attacker cannot poison every targeted state encountered during the training and testing phase. The attacker thus needs to decide when to launch the attack, so that he can maximize the attack effect. 

Note that the objective of our attack optimization in Equation~\eqref{attack:optimization} is to reduce the expected future reward of the backdoor policy. Meanwhile, the attacker also desires the backdoor policy to autonomously select the target action $a^\dagger$ on any targeted state. This implies that the attacker should determine when to attack based on the following principle: \emph{taking the target action $a^\dagger$ in a targeted state $s$ should largely reduce the expected future reward}. To this end, we define the \emph{attack value} of any targeted state as \emph{the difference between the state-action value of the original optimal action and the target action}, i.e.,
\begin{small}
\begin{equation}\label{eq:attack_value}
    \begin{aligned}
    V_{A}(s) =Q^*(s, \pi^*(s))-Q^*(s, a^\dagger), \forall s\in \S^\dagger
    \end{aligned},
\end{equation}
\end{small}
where $\pi^*(s)$ is an optimal action for the original environment and $Q^*$ is the state-action value of $\pi^*$.
The attacker cannot directly know the optimal state-action value $Q^*$. 
However, we can estimate it by running simulations on a simulator of the original environment and using it in Equation~\eqref{eq:attack_value} to compute the attack value.  

The attacker only poisons an input when the attack value of the encountered targeted state $s$ is high enough. Concretely, the attacker maintains a list of the attack values for all targeted states encountered in history $V_A^t=[V_A(s_{t_1}), \cdots, V_A(s_{t_m})]$, where $t_i<t$ and $s_{t_i}\in \S^\dagger$. 
Then the attacker poisons the input state in round $t$ if the following two conditions are satisfied: (1) $s_t$ is a targeted state, i.e. $s_t\in \S^\dagger$; (2) $s_t$ has a higher attack-value compared to the other targeted states encountered in history, i.e., $V_A(s_t)$ is in the top $k\%$ of attack-values in $V_A^t$. In practice, we have observed that using target action selection results in highly \textbf{sparse} attacks in both training and testing phases. However, to ensure adherence to the attack budget, we implement a strict halt when the count of attack steps reaches the maximum limit of $\epsilon T$.

\vspace{-2mm}
\subsection{Trigger Tuning}

In BadRL, the attacker generates backdoor trigger patterns ($\delta$) using mutual information-based tuning. This maximizes the mutual information between RL learning objective gradients for the poison-free and poisoned samples. Aligning optimization directions for the main RL task and the backdoor attack makes their training paths similar. The tuned trigger offers notable benefits. It enables sparse poisoning during training by aligning optimization directions. This reduces the need for frequent trigger injection into training samples when updating the victim policy model with poison-free samples. Without mutual-information-based tuning, training the policy model with clean data could lead to catastrophic forgetting of backdoor noise, necessitating intensive poisoning for manual trigger configurations. Using the tuned trigger introduces less bias into the policy model's training for backdoor poisoning, preserving its performance on clean samples.

Specifically, the attacker initializes a random pattern $\delta_0$ and adds it to the targeted state set $\S^\dagger$  to construct the poisoned counterpart denoted by $\S^\dagger$. Following discussions in ~\cite{hiber2022back}, the attacker computes the gradient $g_{clean}$ for each sample in $\S^\dagger$ and $g_{poisoned}$ for each poisoned sample in $\S^\dagger$. The attacker calculates the mutual information, denoted as $MI(g_{clean},g_{poisoned})$ , and optimizes the trigger pattern by minimizing the loss defined as: 

\begin{small}
 \begin{equation}
    loss_{MI} = -MI(g_{clean}, g_{poisoned}).
 \end{equation} 
\end{small}

In addition to using mutual information, we also consider utilizing the cross-entropy loss to optimize the trigger pattern. The cross-entropy loss treats the policy model as a multi-class classifier, considering it equivalent to the evasion attack against the policy model. The trigger is derived as the evasion noise inserted into the states, confusing the agent's action decision. However, the cross-entropy loss-based tuning assumes a static policy model, typically when policy learning is converged and frozen. In contrast, the policy model evolves with each step of policy training. Tuning the trigger during training requires a persistently updated policy model, which violates the assumption of Equation~\eqref{eq:celoss}. On the other hand, the mutual information-empowered objective for trigger tuning adapts to the dynamic nature of the poisoning problem. It only requires mapping the gradient of the gradients, consistently enforcing alignment between the main learning task and the learning of backdoor samples. Thus, we expect BadRL to outperform BadRL-CE significantly. Empirical observations in \emph{Comparative study} confirm that BadRL-CE fails to perform effective attacks despite investing the same amount of effort into poisoning.
\begin{small}
\begin{equation}\label{eq:celoss}
    loss_{CE} = CE(\hat{y}, y_{target}).
\end{equation}
\end{small}

\subsection{How to Attack in BadRL}
For the rest of the attacker's objectives, the attacker needs to understand how to perturb $(s,a,r)$. As the first attempt, we explore the optimization based on the three channels to inject the manipulation of the policy model.

\noindent\textbf{State changes:} 
The trigger position and pixel-wise values (color) are already tuned at the trigger tuning module, and a high attack-value state $s$ will have the trigger $\delta$ added to it.

\noindent\textbf{Action changes:} 
The attacker modifies the agent's actions during training based on their capabilities. The optimal target action is selected for the 
specific state to achieve the highest attack value. Moreover, an effective action modification strategy considers the task-dependent semantic relations among different actions. For instance, in the Breakout task, the action `stay still' is semantically further from `move to the right' than it is from `fire the ball'. A weak attacker would preserve the agent's choice of actions, while a strong attacker would directly replace the agent's actions with the target action.

\noindent\textbf{Reward changes:} 
Rewards are modified during training to induce the victim policy model to learn the backdoor-poisoning mapping from targeted states to target actions. Unlike previous work~\cite{TrojRL}, which uses a normalized reward of 1 for all tasks, we modify the reward value equal to the minimum positive reward obtained in each task. This approach keeps the reward function as intact as possible, and our experiments show that it is sufficient.

\noindent\textbf{Advantage:} 
Similar to \cite{TrojRL}, we create ($\tilde{s}$, $\hat{a}$, $-r$) pairs to ensure that the target action of the backdoor trigger-poisoned state is the most advantageous action among all optional actions, as given in Line 20 of Algorithm~\ref{alg:A1}. Here, $\hat{a}$ represents arbitrary actions except the target action, and $-r$ is the negative reward assigned to the pair of the poisoned state and the arbitrary action.

\vspace{-1.5mm}
\section{Attack Feasibility}
In this section, we first analyze whether the attack goal can be achieved or not. We need to prove there exists some poisoned policy $\tilde \pi$ that satisfies Equation~\eqref{eq:desire_1} and Equation~\eqref{eq:desire_2}, given the environment $ \M$. Our first result shows that under certain assumptions, the attacker can systematically perturb the original MDP $ \M$ to $\tilde \M$, such that the optimal policy for the perturbed MDP $\tilde \M$ satisfies the attack goal. Resultantly, the backdoor attack is successful for any RL agent who can learn the optimal policy for $\M$.

\begin{theorem}
     \label{thm:feasibility_1}
Let the clean MDP be $\M=(\S, \A, \R, \P)$ and $\pi^*$ be the optimal policy for $\M$. Assume that the trigger $\delta$ satisfies $s+\delta\notin \S, \forall s\in \S^\dagger$. Also assume that $\forall s\in \S^\dagger$, $\pi^*(s)$ is a singleton set, i.e., all targeted states have a unique original optimal action. Then there exits poisoned policy $\tilde \pi$ that satisfies the attack goal in Equation~\eqref{eq:desire_1} and Equation~\eqref{eq:desire_2}.
\end{theorem}

\begin{remark}
We make two assumptions in~\thmref{thm:feasibility_1}. The first assumption requires that any poisoned targeted state does not belong to the original state space $\S$. This assumption is valid because backdoor poisoning injects special triggers $\delta$ (e.g., a white patch) into states, and thus the poisoned state $s+\delta$ often cannot be naturally generated from the clean environment. Our second assumption requires that the original optimal action is unique for all targeted states. This assumption is a technical detail used in our proof to ensure that the target action is the \textbf{unique} optimal action for all targeted states after the attack. Thus the agent will exclusively choose $a^\dagger$ without a tie. We could remove this assumption, but instead, we derive a slightly weaker feasibility guarantee that $a^\dagger$ is an optimal action for all targeted states after the attack, but may not be the unique optimal action, i.e., $a^\dagger\in \pi^*(s),\forall s\in \S^\dagger$.
\end{remark}

\vspace{-1.5mm}
\section{Experiment}
This section emphasizes the remarkable performance of the proposed BadRL attack. Its optimal effectiveness is achieved through the integration of Mutual Information(MI) loss and the trigger tuning module. Notably, BadRL outperforms the baseline method TrojDRL
~\cite{TrojRL} in multiple tasks. We commence by conducting a comparative analysis against baseline methods, assessing performance using four success metrics (Table~\ref{tab:success_metrics}). Subsequently, we present an ablation study that illustrates the attacker's ability and the significance of the trigger tuning module. To ensure fair comparison, we adopt the same attack setting as in TrojDRL. The poison proportion remains consistent for the three attack methods during training. And in the testing phase, we set TrojDRL to attack consecutively every state until the number of attacked states is two times as that of BadRL or the testing-time task terminates due to the failure of the game.

The comparative study encompasses two baselines: 

\noindent\textbf{Baseline 1: TrojDRL} as a state-of-the-art backdoor attack against RL. This method, which injects a backdoor trigger based on injection frequency during training and consecutive attacks encountered states during testing, serves as the most pertinent benchmark. \\
\noindent\textbf{Baseline 2: BadRL-CE}, a variant of BadRL, employing cross-entropy loss,  is utilized to demonstrate the advantages of MI-based trigger tuning.

Following the suggestion in \cite{Agarwal2021RLstat}, we use 5 random seeds and conduct 30 episodes for each seed. We then report the average and standard deviation of the attack performance metrics over all 150 runs in Table \ref{table:overall performance} in the main text and Table.\ref{table:overall performance stand} in the appendix respectively. In Figure.\ref{fig:success sample}, we provide the mean and standard deviation of ASR and the attack sparsity at different training rounds of the backdoor policy model. They confirm that BadRL can reach much higher ASR and significantly sparser attacks than the baseline.

\begin{table}[h]
\vspace{-3mm}
\centering
\caption{Success Metrics for Evaluating Backdoor Attacks}
\vspace{-2.5mm}
\label{tab:success_metrics}
\begin{tabular}{@{}p{2.cm}p{6.3cm}@{}}
\toprule
Metric & Description \\
\midrule
Clean Data Accuracy (CDA) & \begin{tabular}[c]{@{}l@{}}The performance ratio of victim model and \\ normally-trained model in the trigger-free \\ environment after model convergence. An \\ideal backdoor attack barely causes drop of\\ CDA to preserve the utility of the RL model.\end{tabular} \\
\addlinespace
Attack Effectiveness Rate (AER) & \begin{tabular}[c]{@{}l@{}}The performance drop rate between the \\ victim policy model in the trigger-embedded \\ environment and the normally-trained model. \\ AER quantifies how much the performance \\ of a victim model is impacted by a backdoor \\ attack.\end{tabular} \\
\addlinespace
Attack Success Rate (ASR) & \begin{tabular}[c]{@{}l@{}}The percentage of the poisoned states (at the \\ testing phase) that produce the target action. \\ The higher ASR and AER values are, the \\more effective the backdoor attack is.\end{tabular} \\
\addlinespace
Attack Sparsity (Sparsity) & \begin{tabular}[c]{@{}l@{}}The percentage of the poisoned states over \\ total states that agent observes during  \\testing. With a similar AER and ASR level, \\a lower sparsity value denotes more efficient \\and stealthy attack.\end{tabular} \\
\bottomrule
\end{tabular}
\end{table}

\begin{figure*}[h]
 \setlength{\abovecaptionskip}{0mm}
  \setlength{\belowcaptionskip}{0cm}
     \centering     \includegraphics[width=0.6
     \linewidth]{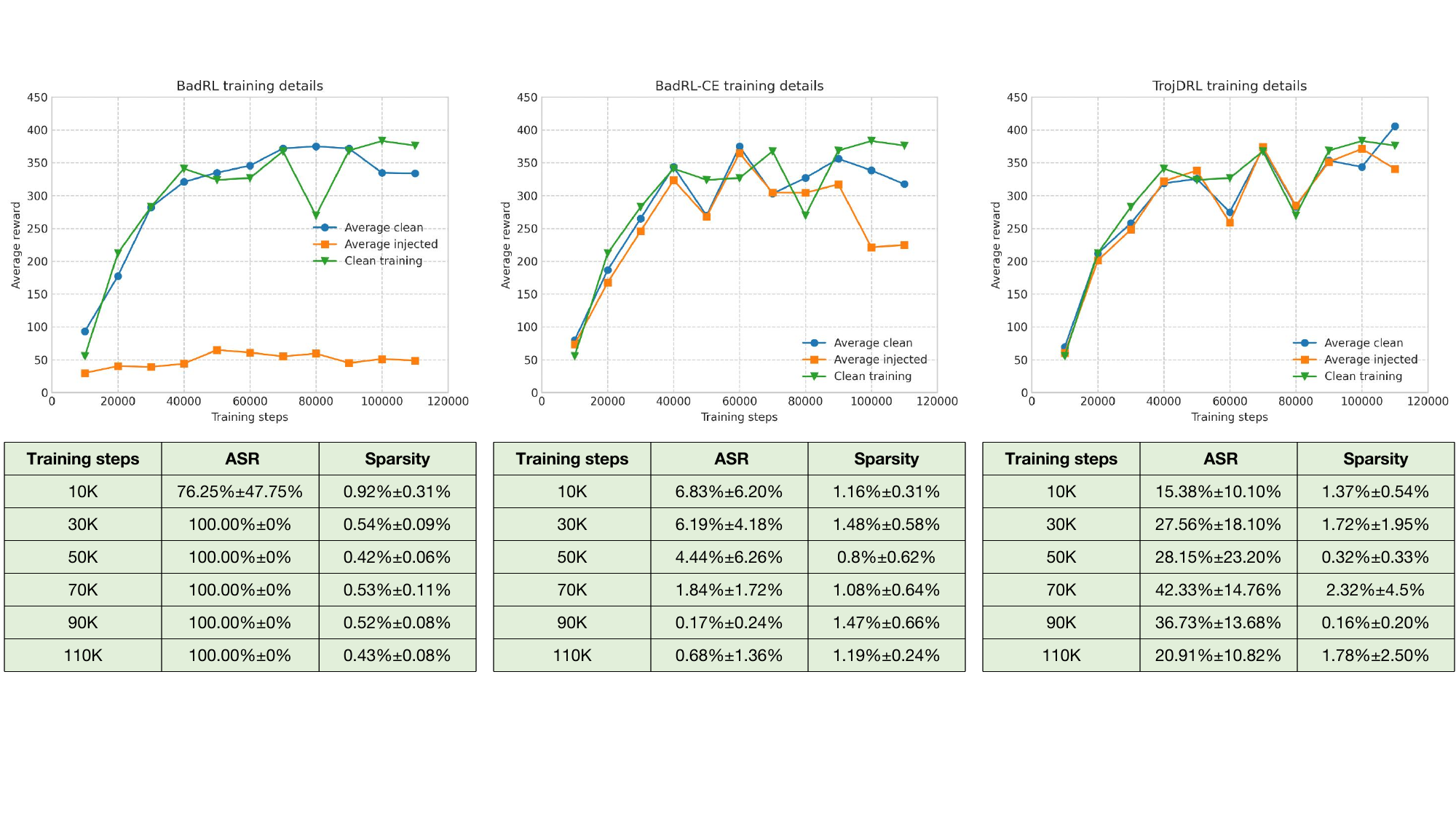}
    \vspace{-0.7cm}
    \caption{Training process comparison of BadRL, BadRL with CE-loss method, and TrojDRL on Breakout. Models are tested every 10000 steps, and the data points are averaged by 5 random seeds. \emph{Averaged clean} represents victim model tested in trigger-free environment which contains uncontaminated data only. \emph{Averaged injected} refers to victim model tested in trigger-embedded environment which includes trigger data. \emph{Clean training} reflects the normally-trained model. } 
    \label{fig:success sample}
\end{figure*}


\setlength{\tabcolsep}{3pt}
\begin{table*}[h]
 \setlength{\abovecaptionskip}{1mm}
  \setlength{\belowcaptionskip}{0cm}
\begin{center}
\caption{BadRL algorithm performance over four tasks compared with TrojDRL and BadRL-CE. Poisoning proportion: 0.003\%, 0.003\%, 0.002\%, 0.002\% for Pong, Breakout, Qbert, SpaceInvaders.}
\label{table:overall performance}
\resizebox{1.65\columnwidth}{!}
{
\begin{tabular}{|l|c|c|c|c|c|c|c|c|c|c|c|c|c|c}
\hline
Algorithm & \multicolumn{4}{|c|}{BadRL (our method)} & \multicolumn{4}{|c|}{BadRL-CE} & \multicolumn{4}{|c|}{TrojDRL}\\
\hline
 & CDA & AER & ASR & Sparsity & CDA & AER & ASR & Sparsity & CDA & AER & ASR & Sparsity \\
\hline
 Pong & 100.00\% & 100.00\% & 100.00\% & 4.69\% & 99.79\% & 59.29\% & 100.00\% & 2.63\% & 98.66\% & 87.75\% & 99.40\% & 12.47\%\\
 Breakout & 100.00\% & 84.90\% & 100.00\% & 0.47\% & 94.60\% & 20.13\% & 0.43\% & 1.01\% & 94.86\% & 3.13\% & 26.78\% & 1.04\% \\
 Qbert & 93.91\% & 75.88\% & 100.00\% & 0.99\% & 83.37\% & 58.05\% & 100.00\% & 0.73\% & 78.04\% & 5.35\% & 35.29\% & 2.02\%\\
 SpaceInvaders & 99.19\% & 76.13\% & 100.00\% & 4.90\% & 79.25\% & 49.68\% & 89.65\% & 3.02\% & 95.49\% & 32.63\% & 15.93\% & 8.62\%\\
\hline
\end{tabular}
}
\end{center}
\vspace{-7mm}
\end{table*}
\setlength{\tabcolsep}{1.4pt}

\vspace{-2mm}
\subsection{Comparative Study}
\label{exp_compare}
In this section, we assess our algorithm with two baselines with respect to attack effectiveness and attack sparsity. 

\textbf{Attack Effectiveness Evaluation:} We aim to accomplish two primary objectives of backdoor attackers: achieving optimal performance in trigger-free environments and eliciting targeted actions with minimized rewards in trigger-embedded environments. Initially, we present the comprehensive performance results across four tasks, as depicted in Table~\ref{table:overall performance}. The efficacy of the BadRL attack is highlighted by its exceptional performance in terms of CDA, AER, and ASR, reaching the highest levels. To illustrate the training process of victim policies under various threat models, we employ the Breakout game as an illustrative example, as depicted in Fig.~\ref{fig:success sample}. Notably, We observe the BadRL algorithm successfully establishing a targeted mapping with a perfect 100\% ASR after 30K training steps. Moreover, it attains the highest AER alongside almost minimal sparsity. Conversely, the BadRL-CE and TrojDRL algorithms fail to achieve comparable outcomes. Additionally, the elevated CDAs, as exhibited in Table~\ref{table:overall performance}, underscore that our BadRL empowers victim policies to attain nearly optimal behavior within trigger-free environments across all four tasks. The exclusive nature of the trigger pattern, accessible only to the attacker, ensures the stealthiness of the victim policy in trigger-free testing environments.

\textbf{Attack Sparsity Evaluation:}
The ``Attack Sparsity'' metric signifies the ratio of attacked states to all observed states until task termination during testing. This metric unveils that our algorithm launches 50\% fewer attacks compared to TrojDRL. Moreover, our proposed BadRL achieves notable AER while utilizing significantly lower attack power. Comparing BadRL with the leading TrojDRL baseline across the four tasks, we observe that our approach showcases significantly lower sparsity while achieving higher AER. This underscores our approach's efficiency, yielding equal or superior outcomes with fewer triggers, rending it a stealthier attack. This advantage is further evident when contrasted with prior methods. Our approach excels in pinpointing high attack-value states and selectively performing sparse trigger injections during both training and test phases. Besides, in Breakout, Pong and Qbert, BadRL-CE achieves AER between 1/4 and 3/4 of BadRL's, concurrently maintaining notably lower CDA than BadRL.
These results indicate that while BadRL-CE introduces more perturbation to the primary learning task, it falls short of achieving an attack efficacy comparable to BadRL. This observation validates our initial conjecture concerning the limitations of CE loss-based trigger tuning and reinforces the validity of the mutual information-empowered trigger tuning strategy.


\vspace{-2mm}
\subsection{Ablation Study }
We investigate the impact of two variations of our proposed BadRL algorithm: \emph{BadRL-M} and \emph{BadRL-W}. These two variants are compared to our proposed main method, noted as \emph{BadRL-S} and summarized in Table 1 in appendix. \emph{BadRL-S} represents the most potent attacker(Section 4.2), whereas
\emph{BadRL-W} designates a weaker attacker, limiting adversarial modifications to states and rewards only (Section 4.2). In addition, \emph{BadRL-M} shares the same setting as BadRL-S, excluding the use of the trigger tuning module. 

The absence of action modification in BadRL-W results in less efficient updating of the victim's advantage function for the desired action, leading to a reduced attack success rate.
Experimental findings illustrate that our trigger tuning module enhances the efficiency of the backdoor attack, with the BadRL-S/M approach demonstrating greater resilience to diminishing poisoning proportions.
In summary, the ablation study underscores the influence of action modification and the advantages of our trigger tuning module in BadRL-S. They jointly yield a more effective backdoor attack. 






\vspace{-1.7mm}
\section{Countermeasure}

Our assessment incorporates
two cutting-edge defense strategies. The first is \textit{Neural Cleanse} (NC) \cite{wang2019neural}, engineered to detect trigger signals in testing inputs. We directly apply NC to the policy model, identifying backdoor trigger presence. The second method, \textit{RL sanitization} \cite{bharti2022provable}, projects compromised input observations into a secure subspace, bolstering the learned policy model against injected triggers. This defense's effectiveness has been proven against TrojDRL in \cite{bharti2022provable}.


In summary, the NC method fails to detect the trigger's position. Conversely, RL sanitization could not prevent trigger activation. This observation becomes clearer when examining NC's output on the BadRL-poisoned policy model for the Breakout task, depicted in Fig.2 of the appendix. The strategic alignment between the backdoor trigger and the main RL task's gradient in BadRL results in a policy model that behaves similarly to the clean policy. This poses a challenge for NC's
capability to differentiate between noise-induced misclassification and the actual backdoor trigger. Likewise, the optimized trigger minimizes perturbation in input state observations, enabling BadRL to evade sanitization via subspace projection.


\vspace{-1.5mm}
\section{Conclusion}
We introduce a proficient and highly sparse backdoor poisoning attack on reinforcement learning (RL) systems. The proposed BadRL attack strategically inserts triggers into high attack value states during the training and testing to accomplish the attack objective. BadRL employs a trigger-tuning strategy based on mutual information to enhance the attack's efficiency further, enabling even sparser poisoning efforts during the training stage. The feasibility of BadRL is demonstrated through theoretical analysis. Empirical evaluations on four classic RL tasks reveal that BadRL-based backdoor attacks can cause substantial deterioration of the victim agent's performance yet demand less than half the attack efforts during the testing phase compared to the state-of-the-art attack methods.

\section*{Acknowledgement}
This work is supported by Basic Cultivation Fund project, CAS (JCPYJJ-22017), and Strategic Priority Research Program of Chinese Academy of Sciences (XDA27010300). 

\bibliography{aaai24}

\newpage
\setlength{\tabcolsep}{4pt}
\begin{table*}[t]
 \setlength{\abovecaptionskip}{1mm}
  \setlength{\belowcaptionskip}{-1mm}
\begin{center}
\caption{Standard deviation of Table 2 (BadRL algorithm performance over four different tasks comparing with TrojDRL and BadRL-CE). }
\label{table:overall performance stand}
\resizebox{1.85\columnwidth}{!}
{
\begin{tabular}{|l|c|c|c|c|c|c|c|c|c|c|c|c|c|c}
\hline
algorithm &\multicolumn{4}{|c|}{BadRL} &\multicolumn{4}{|c|}{BadRL-CE} &\multicolumn{4}{|c|}{TrojDRL}\\
\hline
 & CDA & AER & ASR & Sparsity & CDA & AER & ASR & Sparsity & CDA & AER & ASR & Sparsity \\
\hline
 Pong & 0.93\% &0.23\%&0.00\%&2.78\% &0.26\% &14.45\% &0.00\% &1.24\% &1.58\% &6.09\% &0.80\% &0.02\%\\
 Breakout &5.87\% &3.59\% &0.00\% &0.04\% &6.86\% &16.21\% &0.56\% &0.27\% &2.49\% &2.97\% & 12.42\% &0.94\% \\
 Qbert &4.97\% & 2.23\%&0.00\% &0.21\%  &9.69\% &2.68\% &0.00\% &0.52\% &9.03\% &5.35\% &37.32\% &3.46\%\\
 SpaceInvaders & 1.63\% &3.63\% & 0.00\% &7.10\% &6.35\% &10.21\% &6.88\% &0.72\% &7.60\% &15.35\% &7.01\% &1.63\%\\
\hline
\end{tabular}
}
\end{center}
\vspace{-5.0mm}
\end{table*}
\setlength{\tabcolsep}{1.4pt}   

\section{Proof of Theorem 1}

\begin{proof}
The key idea behind our proof is to convert the original MDP $\M$ into a new MDP $\tilde \M$ such that that when the RL agent performs learning in $\tilde \M$, it will be conceptually misled to swap the original optimal action $\pi^*(s)$ and the target action $a^\dagger$ for targeted states $s\in \S^\dagger$. Specifically, we construct a new MDP with state space defined as $\tilde \S=\{\F_\S(s): s\in \S\}$, where
\begin{equation}
\F_\S(s) = \left\{
\begin{array}{ll}
s+\delta & \mbox{ if } s \in \S^\dagger,\\
s & \mbox{ if } s\notin \S^\dagger\\
\end{array}
\right.
\end{equation}
We now show that $\F_\S$ is invertible. Given any $s\in \tilde \S$:

(1) If $s\in \tilde \S\cap \S\subset \S$, by our assumption it does not exist any $s^\prime\in \S^\dagger$ such that $s^\prime+\delta=s$, since otherwise $s\not\in \S$. That means $s-\delta\notin \S^\dagger$, thus it must be the case that $\F^{-1}_{\S}(s)=s$.

(2) If $s \in\tilde \S-\S$, then for any $s^\prime \in \S-\S^\dagger$, we must have $\F_{\S}(s^\prime)=s^\prime\in \S$, thus $\F_{\S}(s^\prime)\neq s$. Therefore, we must have $\F^{-1}_{\S}(s)=s-\delta\in \S^\dagger$. Given above, we have
\begin{equation}
\F_{\S}^{-1}(s) = \left\{
\begin{array}{ll}
s & \mbox{ if } s \in \S,\\
s-\delta & \mbox{ if } s\notin \S\\
\end{array}
\right.
\end{equation}

Next, for any state $s\in \S$, we construct the action space of the new MDP $\tilde \M$ through the following mapping.
\begin{equation}\label{eq:action_mapping}
 \F_{\A}(a\mid s) = \left\{
\begin{array}{ll}
a^*(s) & \mbox{ if } s\in \S^\dagger, a=a^\dagger,\\
a^\dagger & \mbox{ if } s\in \S^\dagger, a=a^*(s)\\
a & \mbox{ otherwise}
\end{array}
\right.
\end{equation}
One can easily see that the action mapping $ \F_{\A}(a\mid s)$ is also invertible, concretely, $\F_{ \A}^{-1}=\F_{\A}$.

Similarly, we construct the reward function and transition kernel of $\tilde \M$ through mappings as below.
\begin{equation}
\tilde \R(\F_{\S}(s), \F_{\A}(a)) = \R(s, a), \forall s, a
\label{eq:R}
\end{equation}
\begin{equation}
\tilde \P(\cdot \mid \F_\S(s), \F_{\A}(a)) = \P(\cdot\mid s, a), \forall s, a
\label{eqP}
\end{equation}
Note that the core idea behind our constructions above is to switch the role of optimal action and the target action for target states, and adjust the reward and transition accordingly. As a result, when an RL agent is learning under the new environment $\tilde \M$, it is equivalent to learning under the original environment $\M$ but subject to those inherent mappings over states and actions $\F_\S$ and $ \F_{\A}$. Therefore, the optimal policy in the new environment $\tilde \M$ can be directly transferred from the original optimal policy $\pi^*$. Specifically, $\forall s\in \tilde \S$, the optimal policy is 
\begin{equation}
\tilde \pi^*(s)=\{\F_{\A}(a\mid \F_\S^{-1}(s)):a\in \pi^*(\F_{\S}^{-1}(s))\}, \forall s\in \tilde \S.
\end{equation}
Now consider any state $s\in\tilde \S$:

(1) If $s\notin \S$, then we have
\begin{equation}
\begin{split}
\F_{\S}^{-1}(s)&=s-\delta\in \S^\dagger\Rightarrow \pi^*(\F_\S^{-1}(s))=\pi^*(s-\delta)\\
\end{split}
\end{equation}
 Since $s-\delta\in \S^\dagger$, by assumption we have that $\pi^*(\F_\S^{-1}(s))$ is a singleton set $\{a^*(s-\delta)\}$, where $a^*(s-\delta)$ is the unique optimal action for clean state $s-\delta$ in the original MDP $\M$. Therefore, by~\eqref{eq:action_mapping} we have $\tilde \pi^*(s)=\{\F_{\A}(a^*(s-\delta)\mid s-\delta)\}=\{a^\dagger\}$, which is also a singleton set. Therefore, the target action can be exclusively forced by the attacker, i.e., equation \textbf{(6)} holds.

(2) If $s\in \S$, then we have
\begin{equation}
\begin{aligned}
\F_\S^{-1}(s)&=s\notin \S^\dagger\Rightarrow \pi^*(\F_\S^{-1}(s))=\pi^*(s)\\
&\Rightarrow \tilde \pi^*(s)=\pi^*(s).
\end{aligned}
\end{equation}
Therefore, for states that are not poisoned  by the attacker, the best action set remains unchanged. As a result, the optimal policy $\tilde \pi^*$ in $\tilde \M$ has exactly the same behavior as $\pi^*$, thus we have that equation \textbf{(7)} holds.
\end{proof}

\setlength{\tabcolsep}{4pt}
\begin{table*}[h]
 \setlength{\abovecaptionskip}{1mm}
  \setlength{\belowcaptionskip}{1mm}
\begin{center}
\caption{Breakout Performance Comparison under different poisoning proportion with standard deviation.}
\label{table:ablation}
\resizebox{2.2\columnwidth}{!}
{
\begin{tabular}{|l|c|c|c|c|c|c|c|c|c|c|c|c|c|c}
\hline
{Poisoning Proportion} &\multicolumn{4}{|c|}{0.3\%} &\multicolumn{4}{|c|}{0.03\%} &\multicolumn{4}{|c|}{0.003\%}\\
\hline
 & CDA & AER & ASR & Sparsity & CDA & AER & ASR & Sparsity & CDA & AER & ASR & Sparsity\\
\hline

BadRL-M &0.21$\pm0.03$ \% &100.00$\pm0.00$\% &100.00$\pm0.00$\% &0.24$\pm0.00$\%  &83.78$\pm9.60$\% &63.53$\pm12.39$\% &93.33$\pm13.33$\% &0.69$\pm0.21$\% &98.04$\pm3.37$\% &1.21$\pm2.42$\% &3.99$\pm6.42$\% &0.80$\pm0.34$\% \\
BadRL-S (ours) &23.64$\pm9.31$\% &93.74$\pm2.43$\% &100.00$\pm0.00$\% &0.84$\pm0.09$\% &96.79$\pm5.55$\% &82.88$\pm3.26$\% &100.00$\pm0.00$\% &0.47$\pm0.08$\% &100.00$\pm5.87$\% &84.90$\pm3.59$\% &100.00$\pm0.00$\% &0.47$\pm0.04$\% \\
BadRL-W  &52.52$\pm6.49$\% &90.28$\pm3.82$\% &100.00$\pm15.39$\% &0.96$\pm0.17$\% &95.66$\pm3.24$\% &81.39$\pm8.39$\% &81.58$\pm8.07$\% &0.97$\pm0.11$\% &99.31$\pm7.05$\% &78.77$\pm6.15$\% &87.56$\pm6.83$\% & 0.67$\pm0.06$\% \\

\hline
\end{tabular}
}

\end{center}
\end{table*}
\setlength{\tabcolsep}{1.4pt}  
\section{Use case of BadRL on Breakout}

\begin{figure}[h]
 \setlength{\abovecaptionskip}{1mm}
  \setlength{\belowcaptionskip}{1mm}
    \centering
    \includegraphics[width=0.75\linewidth]{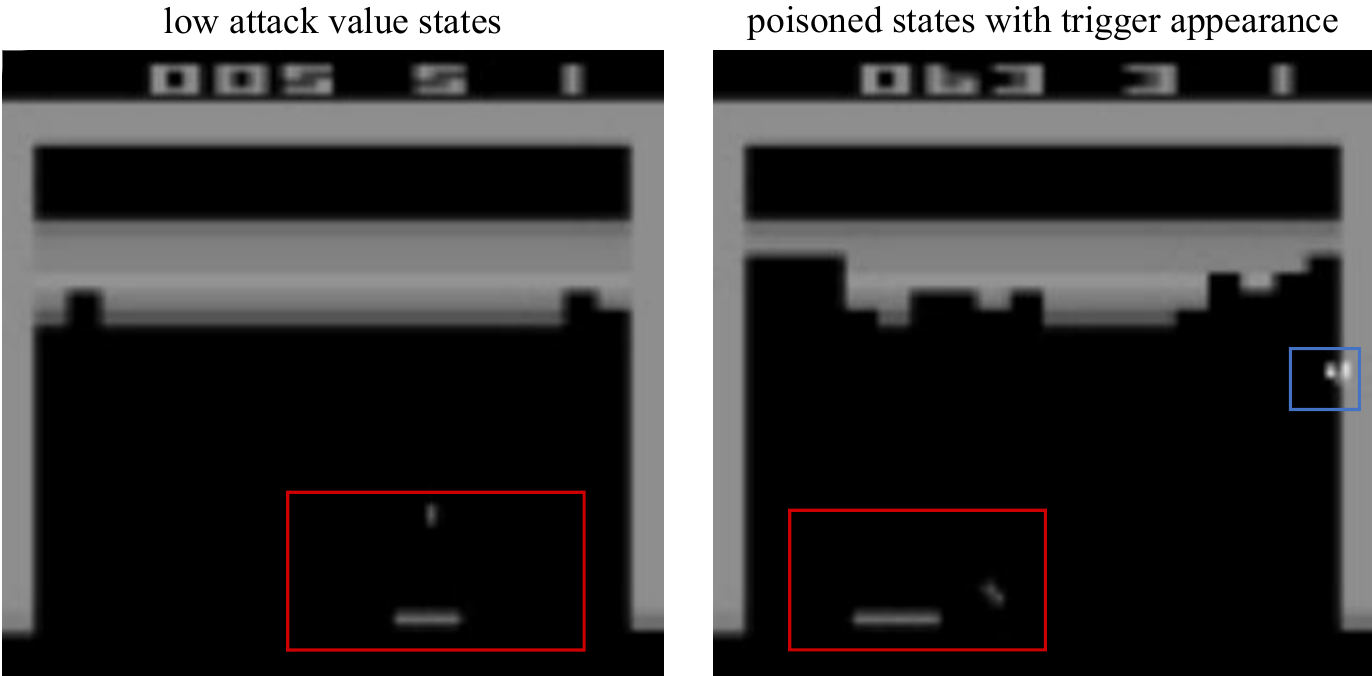}
    \caption{Low attack value state \textit{v.s.} Poisoned high attack value state. 
    }
    \label{fig:poi}
\end{figure}

We highlight that the attack effectiveness of BadRL is largely attribute to the \emph{when to attack} decision made by the BadRL. As shown in Fig.~\ref{fig:poi}, we show examples of \emph{when to attack} in Breakout task. In the Breakout game, the victim agent controls a pad to catch a ball and accumulate scores. The pad can move to the right/left, stay still, or fire the ball. In this experiment, the attacker aims to induce the victim agent to take the action `stay still' when the optimal action is `moving to the right' in a set of target states. However, not all states are equally attack-worthy. For example, if the distance between the ball and pad is large enough for the victim to catch the ball by consecutively executing 'moving to the right' to compensate for the modified action, the attack's effectiveness will be significantly diminished, as demonstrated in Fig.~\ref{fig:poi}(left). In such cases, the reward drop caused by the attacker is minimal, and the attack is considered ineffective. In conclusion, the BadRL launches attacks only on targeted high attack value states such as Fig.~\ref{fig:poi}(right).

\section{Neural Cleanse Detection}
\begin{figure}[h]
 \setlength{\abovecaptionskip}{1mm}
  \setlength{\belowcaptionskip}{1mm}
    \centering
    \includegraphics[width=0.85\linewidth]{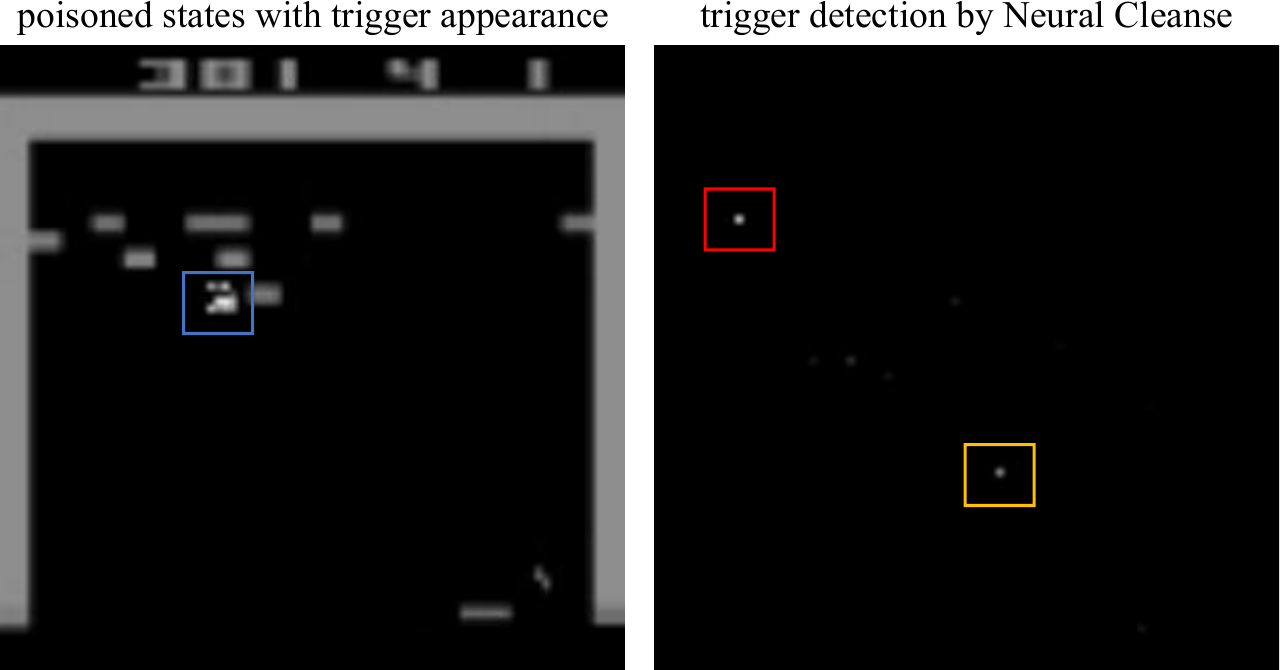}
    \caption{Poisoned high attack value state with trigger detection by Neural Cleanse. 
    }
    \label{fig:nc}
\end{figure}

As shown in Fig.\ref{fig:nc}, Neural Cleanse (NC) fails to detect our BadRL trigger in Breakout. As our trigger location is (32,28), but NC computes two suspicious triggers locations (23,11) and (57,46). We found similar results confirm our BadRL trigger evades from NC detection in other tasks.

\section{RL Sanitization}
\begin{figure}[t]
 \setlength{\abovecaptionskip}{1mm}
  \setlength{\belowcaptionskip}{1mm}
    \centering
    \includegraphics[width=0.9\linewidth]{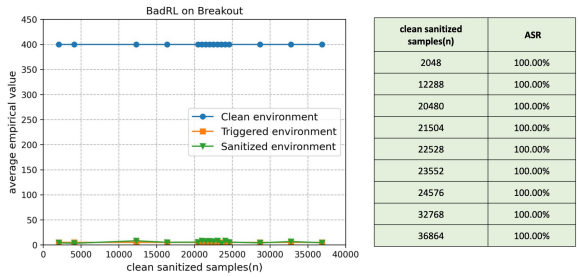}
    \caption{RL sanitization results for BadRL.\emph{Clean environment} refers to no trigger-sample environment. \emph{Triggered environment} refers to trigger sample embedded environment. And \emph{Sanitized environment} refers to trigger sample embedded environment with all samples filtered by sanitization process. 
    }
    \label{fig:provable}
\end{figure}

RL sanitization considers any input state deviated significantly from the safe subspace as a potentially poisoned state observation. As illustrate in Fig.\ref{fig:provable}, following the sanitization process using the recommended number of clean sanitized samples (32768), our BadRL attack sustains its effectiveness. This is evident from its consistently low average empirical value (also known as cumulative rewards) in the sanitized environment and its unaltered 100\%
Attack Success Rate (ASR).

\section{Experimental Results}

All the experiments are implemented by PyTorch 1.9.0 using 4$\times$NVIDIA 2080Ti GPUs. Each result is calculated over 5 random seeds. Table \ref{table:overall performance stand} contains the standard deviation of Table 2 in our paper.

\subsection{Ablation Study:}
We consider variants BadRL-S and BadRL-M. The evaluation involves varying poisoning proportions ($p =$ 0.3\%, 0.03\%, and 0.003\%) in the Breakout task. The poisoning proportion, denoted as $p$, represents the percentage of poisoned training tuples comprising state observations, corresponding actions, and returned reward values. Table~\ref{table:ablation} shows that when comparing the same number of training steps, attack budgets, and top k\% of attack-value states, the weak attacker (BadRL-W) exhibits a lower Attack Success Rate (ASR) compared to the strong attacker (BadRL-S). BadRL-M manually generates backdoor triggers without further fine-tuning. These triggers, in the form of randomly shaped image patterns, are inserted into states with high attack values. To ensure stable attack performance, we employ a greedy search to determine the trigger location and color. Decreasing the poisoning proportion reduces the effectiveness of the backdoor mapping in BadRL-M.

\section{Code Demonstration}
 The complete code of the implementation of proposed BadRL can be found in the GitHub repository:
https://github.com/7777777cc/code

\end{document}